\documentclass[runningheads]{llncs}
\usepackage{graphicx}

\usepackage{tikz}
\usepackage{comment} 
\usepackage{amsmath,amssymb} 
\usepackage{color}

\begin{document}
\pagestyle{headings}
\mainmatter
\def\ECCVSubNumber{100}  

\title{Robust RGB-based 6-DoF Pose Estimation without Real Pose Annotations} 

\titlerunning{Abbreviated paper title}
%
\author{Zhigang Li \inst{1} \and Yinlin Hu \inst{2} \and Mathieu Salzmann \inst{2,3} \and Xiangyang Ji \inst{1}}
\authorrunning{Z. Li, Y. Hu, M. Salzmann and X. Ji}
%
\institute{Tsinghua University, Beijing, China \\
\email{\{lzg15, xyji\}@mails.tsinghua.edu.cn}\\
\and EPFL CVLab, Lausanne, Switzerland \\
\email{\{yinlin.hu,mathieu.salzmann\}@epfl.ch}\\
\and ClearSpace, Lausanne, Switzerland \\
\email{mathieu.salzmann@clearspace.today}
}
\maketitle

\begin{abstract}
While much progress has been made in 6-DoF object pose estimation from a single RGB image, the current leading approaches heavily rely on real-annotation data. As such, they remain sensitive to severe occlusions, because covering all possible occlusions with annotated data is intractable. In this paper, we introduce an approach to robustly and accurately estimate the 6-DoF pose in challenging conditions and without using any real pose annotations. To this end, we leverage the intuition that the poses predicted by a network from an image and from its counterpart synthetically altered to mimic occlusion should be consistent, and translate this to a self-supervised loss function. 
Our experiments on LINEMOD, Occluded-LINEMOD, YCB and new Randomization LINEMOD dataset evidence the robustness of our  approach. We achieve state of the art performance on LINEMOD, and Occluded-LINEMOD in without real-pose setting, even outperforming methods that rely on real annotations during training on Occluded-LINEMOD.
\end{abstract}

\section{Introduction}
Accurately estimating the rotation and translation of a 3D object model relative to the camera from a single RGB image, referred to as 6-DoF pose estimation, has many real-world applications, such as augmented reality, mobile robotics, and autonomous navigation. 
As such, it has attracted continuous attention in the research community. 
Traditionally, this task was tackled as a geometric problem, solved by matching 2D image features with 3D object keypoints~\cite{haralick1989pose,unsalan2007model,aubry2014seeing}.
While effective for well-textured objects, these methods do not generalize to the poorly-textured case. Therefore, recent advances in the field have focused on a deep-learning-based approach~\cite{kehl2017ssd,xiang2018posecnn,AAE_2018_ECCV,pavlakos20176,rad2017bb8,tekin18_yolo6d,hu2019segmentation,peng2019pvnet,li2019cdpn,less_is_more,zakharov2019dpod,wang2019NOCS,park2019pix2pose}. 

In this context, initial attempts were made to infer the pose directly from the image~\cite{kehl2017ssd,xiang2018posecnn,AAE_2018_ECCV,kendall2015posenet}. 
Noticing that the mapping from image space to rotation space, i.e., $SO(3)$, was difficult for deep networks to model accurately, recent methods typically draw inspiration from traditional ones and use deep networks to establish 2D-3D correspondences, from which the pose is obtained using a Perspective-n-Point (PnP) algorithm. To construct the correspondences, the network is trained to either detect pre-defined object keypoints~\cite{rad2017bb8,tekin18_yolo6d,hu2019segmentation,peng2019pvnet}, or regress 3D object coordinates from the image~\cite{li2019cdpn,less_is_more,zakharov2019dpod,wang2019NOCS,park2019pix2pose}.
In any event, while these methods achieve impressive results, they rely on large amounts of annotated 
real images, which are time-consuming to obtain~\cite{kaskman2019homebreweddb}.
As a consequence, these techniques remain sensitive to severe occlusions, because covering the space of all possible occlusions with real images is intractable.

A natural way to palliate for the lack of data consists of performing data augmentation. In the context of 6-DoF pose estimation, this is commonly referred to as
Domain Randomization (DR)~\cite{kehl2017ssd,AAE_2018_ECCV,zakharov2019dpod}. Concretely, the data is complemented with semi-realistic synthetic images, where the 3D object model is rendered on a real background, followed by diverse augmentation techniques, such as varying lighting conditions, contrast, blur, and occlusion by removing small image blocks. While DR was indeed shown to improve the pose estimation accuracy, its benefits on the final, real test images remain limited, in large part because existing DR strategies do not tackle the problem of severe occlusions, and thus fail to address one of the main challenges in pose estimation. As such, relying solely on DR yields results that are far from matching the state of the art~\cite{kehl2017ssd,AAE_2018_ECCV,zakharov2019dpod}, and most methods still require access to annotated real images, which limits their applicability.

In this paper, we introduce a robust approach to 6-DoF pose estimation that jointly tackles the problems of severe occlusions and of lack of annotated real images.
To this end, we propose to leverage easily-obtainable real object images \emph{without} pose annotations to help the model to handle challenging real-world situations. Specifically, we design a new truncation DR strategy that mimics the presence of large occlusions, and, as depicted by Fig.~\ref{fig:sspn}, encourage 6-DoF consistency between the pose estimates obtained from the unlabeled real data with and without DR. This, in conjunction with synthetic DR data, for which we have annotations, allows us to bridge the domain gap, so as to obtain a network that performs well in the real world, even in challenging conditions. Our approach is versatile and generalizes to diverse pose estimation backbones.

\begin{figure}[tb]
\centering
\includegraphics[width=12cm]{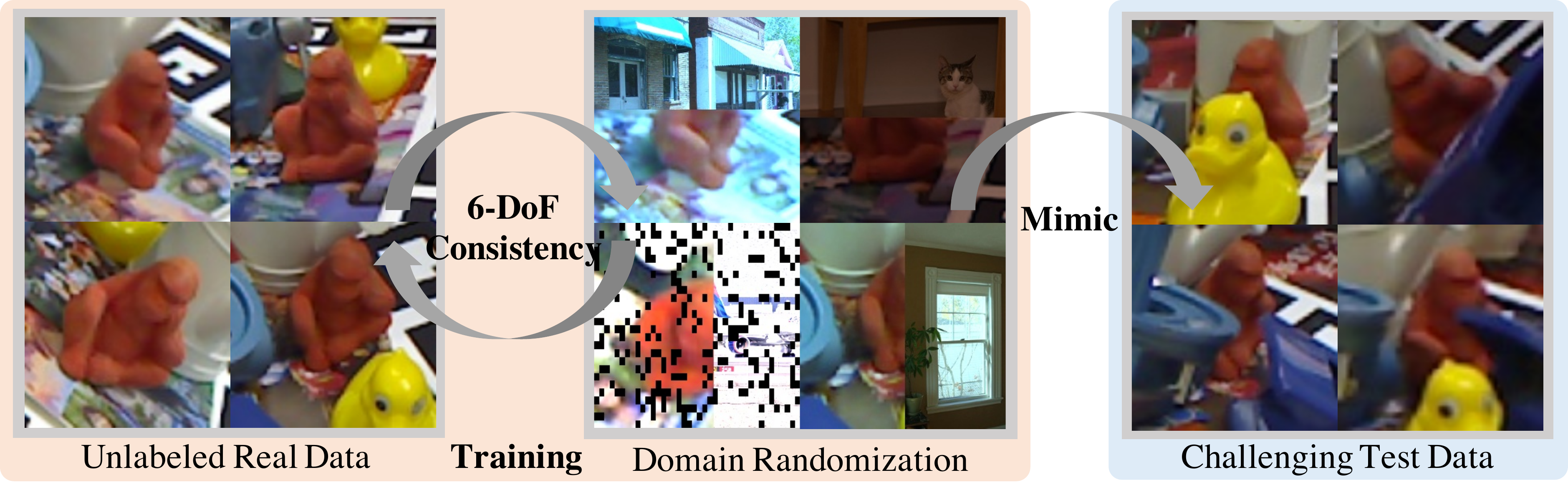}
\vspace{-0.5em}
\caption{Existing pose estimation methods struggle in the presence of severe occlusions and the absence of real annotated data. In this work, 
we exploit 6-DoF consistency between an input image and its DR augmented counterpart to improve the network's robustness and accuracy in challenging situations, without requiring real annotations.}
\label{fig:sspn}
\vspace{-1em}
\end{figure}

Our contributions can thus be summarized as follows:
\begin{itemize}
\item We introduce a truncation DR strategy that explicitly addresses the challenging problem of severe occlusions in 6-DoF pose estimation.
\item We develop a self-supervised learning approach for 6-DoF pose estimation, which can be used to train a network without real 6-DoF annotations.
\item We design a Self-supervised Siamese Pose Network (SSPN) for self-supervised coordinates-based pose estimation, which accurately predicts the object pose even in the presence of severe occlusions.
\item We introduce a Randomization LINEMOD (Rand-LINEMOD) dataset to evaluate the robustness of pose estimation in the real domain. Our results show that domain randomization does not suffice to provide robustness to some real-world challenges (e.g., severe occlusion), which explains why existing DR-based approaches only yield limited performance in these cases.
\end{itemize}

Our comprehensive experiments on several large datasets highlight the benefits of our approach. On the LINEMOD dataset, our approach significantly outperforms the competitors that also do not use real pose annotations. On the Occluded LINEMOD dataset, we achieve the state-of-the-art performance, even outperforming methods that rely on real annotations during training. On the YCB dataset, our approach generally improves the performance on all metrics. 

\section{Related Work}

\subsection{RGB-based 6-DoF Pose Estimation}
Traditional approaches estimate the object pose by establishing correspondences between the image and the object model using hand-crafted features. While effective in some scenarios, these methods are sensitive to cluttered background and varying illumination conditions, and cannot handle poorly-textured objects.
Recently, CNN-based approaches have therefore emerged as an effective alternative, achieving remarkable performance in this field.
Some initial methods~\cite{xiang2018posecnn,kendall2015posenet} utilize the network in an end-to-end fashion to directly regress the pose from the image. 
While these techniques generalize well to diverse objects and are fast at inference, their performance remains limited~\cite{sattler2019understanding}, requiring a post-processing refinement, such as ICP in the presence of depth.
As a consequence, many works adopt an indirect strategy, consisting of first establishing 2D-3D correspondences, which can then be used to estimate the pose. Such correspondences can be obtained by detecting pre-defined object keypoints in the image. In~\cite{peng2019pvnet,pavlakos20176} the keypoints are defined as semantic object parts. However, the diversity of such semantic keypoints across objects restricts the generalizability of this approach. By contrast,~\cite{rad2017bb8,tekin18_yolo6d,hu2019segmentation} exploit the corners of the 3D object bounding box in the image as keypoints. While effective, the deviation of the corners from the object makes the task challenging. 
Another approach to obtaining 2D-3D correspondences consists of regressing the 3D coordinates of each object pixel. This strategy yields dense correspondences and has proven to be robust to occlusion. However, it requires using a time-consuming RANSAC-based algorithm at test time, which slows down the inference process.
In any event, while the current state-of-the-art methods achieve remarkable performance, they rely heavily on annotated real training data, and their accuracy in the presence of severe occlusions remains unsatisfying.
In this work, we aim to robustly and accurately estimate the pose without requiring any real-world pose annotations.

\subsection{Domain Randomization (DR)}
To palliate for the lack of annotated real images, one typically relies on synthetic data, generated by exploiting the availability of 3D models for the objects of interest. In this context, DR is usually adopted to help bridge the domain gap between synthetic and real images, and improve robustness. It refers to augmenting the synthetic images with diverse modifications, such as varying background, lighting conditions and noise, so as to yield semi-realistic images that help the model generalize to the real domain~\cite{AAE_2018_ECCV}. DR is widely used in pose estimation.
For example, in~\cite{kehl2017ssd,zakharov2019dpod,li2019cdpn,li2018deepim}, the objects are rendered with real images (e.g., MS COCO~\cite{lin2014microsoft}) as background. To improve realism, in~\cite{wang2019NOCS}, a context-aware mixed reality approach is used to render synthetic objects in real tabletop scenes by detecting planar surfaces. While background randomization helps to bridge the synthetic-real domain gap, it does not improve the robustness to other phenomena, such as occlusion and blur. To address this, in~\cite{AAE_2018_ECCV,park2019pix2pose,zakharov2019dpod}, additional augmentation schemes, such as varying brightness, contrast, saturation, blur, and removing small image blocks, were deployed during training to extract augmentation-invariant features. These schemes, however, do not address the problem of severe occlusions, which we do here. Furthermore, while DR indeed improves generalization to the real domain, it remains insufficient on its own for existing approaches to yield robust pose estimates in challenging conditions. In this paper, we analyze the limitations of DR and propose to complement it with unsupervised real-world data to improve the model's robustness.

\subsection{Self-supervised Learning}
Self-supervised learning has achieved tremendous success in computer vision~\cite{jenni2018self,misra2019self,lee2017unsupervised,zhang2016colorful,noroozi2016unsupervised,pathak2016context}. It endows the model with the ability to learn from the data itself. This is achieved by designing a proxy task for which the ground truth can be automatically distilled from the input. For instance, the training of autoencoders is self-supervised since the ground truth is the input itself. When it comes to 6-DoF pose estimation, however, extracting supervisory signal that reflects the final goal from the input is challenging. Recently,~\cite{deng2019self} introduced a self-supervised learning approach for a pose-related task, relying on the assistance of a calibrated robotics system to achieve life-long learning. Unfortunately, such a requirement significantly restricts the applicability of this algorithm. 
Here, we develop a self-supervised learning approach for 6-DoF pose estimation without the need for extra equipment. As discussed below, we achieve this by exploiting 6-DoF consistency in unlabeled real data as supervision. 

\begin{figure}[tb]
\centering
\includegraphics[width=12cm]{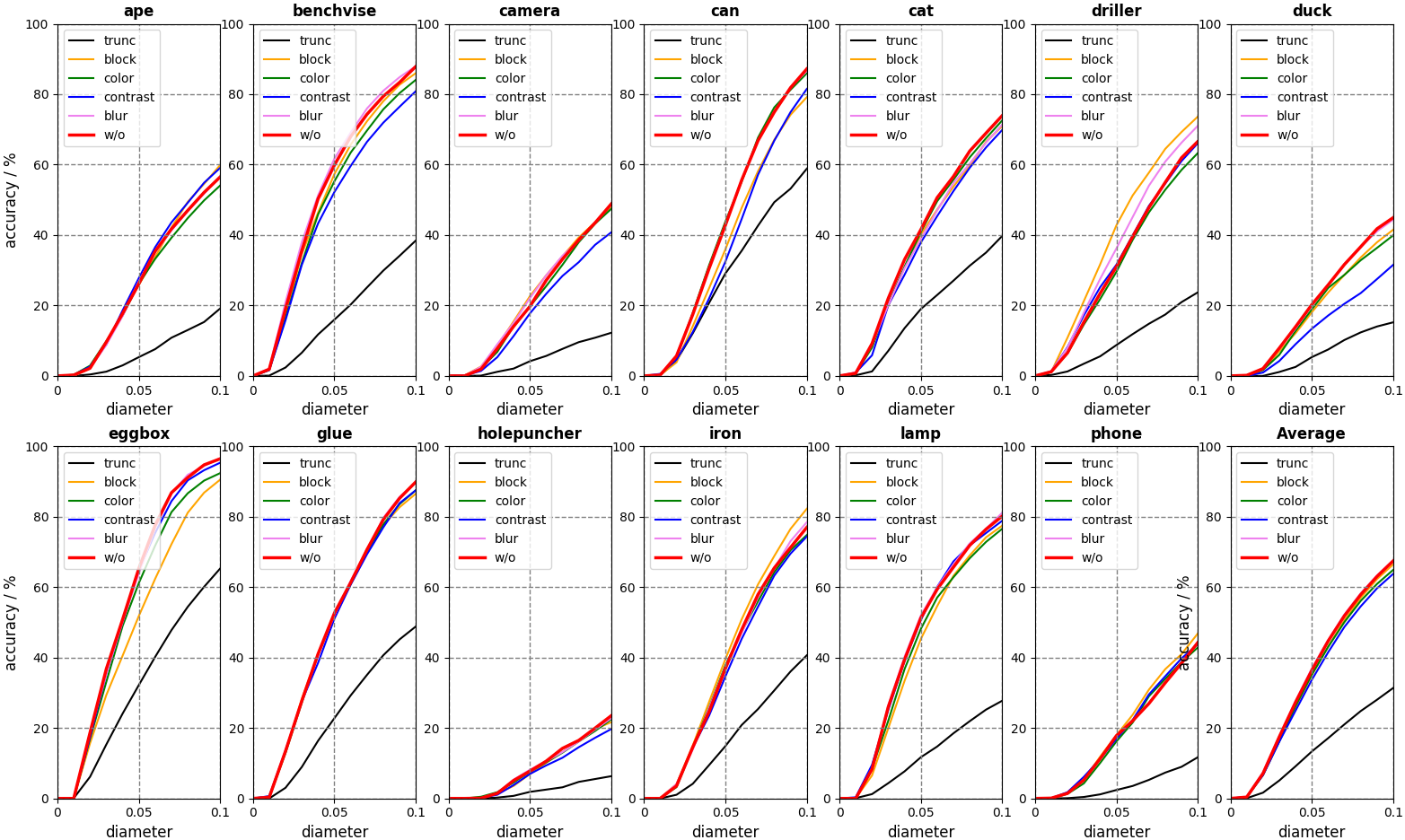}
\vspace{-0.5em}
\caption{Evaluation of the robustness to various real-world disturbances of a network trained on synthetic data only (Metric: ADD; w/o: Baseline results obtained from the LINEMOD test set. The other curves correspond to our Rand-LINEMOD dataset.)} 
\vspace{-1em}
\label{fig:add_rst_no_ssv}
\end{figure}

\section{Domain Randomization (DR) Analysis}
\label{sec:DR}
Let us first analyze the effectiveness of DR, which we will then exploit to develop our self-supervised 6-DoF pose estimation approach. DR is widely used in 6-DoF pose estimation because it was observed to yield a significant performance boost. Its effectiveness is usually attributed to its domain transfer ability, i.e., DR alters the synthetic data so as to approximate the real one. Interestingly, however, DR not only helps to bridge the synthetic-real domain gap; it further improves the model's generalizability in the real domain itself. For instance, adding varying lighting to the synthetic images not only makes them look more realistic but also improves the robustness to diverse illumination conditions in the real world. However, as will be revealed by our analysis below, this endowed robustness varies drastically across different disturbances.

To investigate and evidence this, we introduce the Randomization LINEMOD (Rand-LINEMOD) dataset. Rand-LINEMOD consists of several test sequences, each corresponding to one specific disturbance. In particular, we consider the following standard operations: blur, contrast, varying lighting, and block (small occlusions obtained by removing small image patches). Because none of these operations tackles the problem of severe occlusions, we introduce a new DR strategy named \emph{truncation}. Specifically, this operation consists of removing large portions of the object by sampling a center point from a truncated distribution with a random cutting direction. During training, the removed area is filled with random real-world images (e.g. PASCAL VOC2012\cite{VOC}). We constructed Rand-LINEMOD by introducing such disturbances to the test sequences of the LINEMOD dataset. This therefore allows us to evaluate the robustness of a model in the real domain by measuring the performance discrepancy between the LINEMOD and Rand-LINEMOD datasets.

We evaluate the coordinate-based approach that will constitute the backbone of our self-supervised method\footnote{The details of the coordinate-based method are provided in Section~\ref{Sec.Details}.} on the Rand-LINEMOD dataset. The model was trained on synthetic images with diverse augmentations (blur, contrast, varying lighting, block and truncation) to improve robustness. After training, we evaluate its performance according to the average 3D distance between the model points transformed using the predicted pose and the ground-truth one. Specifically, we report the accuracy obtained by thresholding this distance, commonly referred to as the ADD metric, and plot curves obtained by varying this threshold for different disturbances and different objects in Fig.~\ref{fig:add_rst_no_ssv}, where the performance on the LINEMOD dataset acts as a baseline. On several  Rand-LINEMOD sequences (blur, contrast, varying lighting, block), the model yields similar performance to that on LINEMOD, which evidences the robustness brought about by domain randomization to some disturbances in the real domain. However, for severe occlusions (i.e., our truncation operation), the performance drops significantly for all objects. 
This analysis shows that, for existing methods, DR is not sufficient to achieve high robustness in the real domain to some disturbances such as severe occlusions. 
Furthermore, since the model achieves much better performance on ideal test samples (i.e., LINEMOD test sequences) than in challenging conditions (e.g., Rand-LINEMOD truncation sequences), utilizing the easy samples to improve the model's accuracy on the hard ones appears as a natural way forward. Below, we rely on this intuition to introduce our self-supervised 6-DoF pose estimation framework.

\section{Method}

Let us now present our approach to self-supervised pose estimation. To this end, we first discuss the notion of 6-DoF consistency in the context of DR, and then introduce our complete framework.

\subsection{6-DoF Consistency in Domain Randomization}

Let $f_\theta$ be a trainable model with parameters $\theta$. Our goal is to utilize real images without pose annotations to improve the model's robustness in the real domain. As evidenced in the previous section, given an input image $X$ for which the model yields an accurate prediction, applying some DR disturbances to $X$ may degrade the pose estimate. However, the predictions for $X$ and the transformed $X$ should be consistent. This is what we propose to enforce here.

Specifically, the operations $\mathcal{A}$ in DR can be categorized into two groups:  I. the pose-relevant operations $\mathcal{A}_r$ (rotate, translate, zoom in/out, etc.); and II. the pose-irrelevant operations $\mathcal{A}_i$ (blur, colorization, noise, occlusion, truncation, etc.).  The operations in $\mathcal{A}_i$ preserve the pose, while those in $\mathcal{A}_r$ do not. 

Consistency can thus be expressed by the fact that, for $\mathcal{A}_i$, we would like to satisfy the constraint that $f_\theta(X)$ should be equal to $f_\theta(\mathcal{A}_i(X))$.
By contrast, for $\mathcal{A}_r$, we have that $\mathcal{A}_r(f_\theta(X))$ should equal $f_\theta(\mathcal{A}_r(X))$.
By grouping these two constraints, and assuming known transformations $\mathcal{A}_i$ and $\mathcal{A}_r$, 
we introduce a self-supervised loss function expressed as
\begin{align}\label{Eq:s}
 \mathcal{S} = \Vert f_\theta(\mathcal{A}_i(\mathcal{A}_r(X))) - \mathcal{A}_r(f_\theta(X)) \Vert\;.
\end{align}
In other words, we exploit the prediction of the network on the easy sample $X$ as a pseudo ground truth for the more challenging sample obtained by DR. As will be shown by our experiments, this suffices to significantly boost the robustness of the network in the real domain.

\begin{figure}[tb]
\centering
\includegraphics[width=12cm]{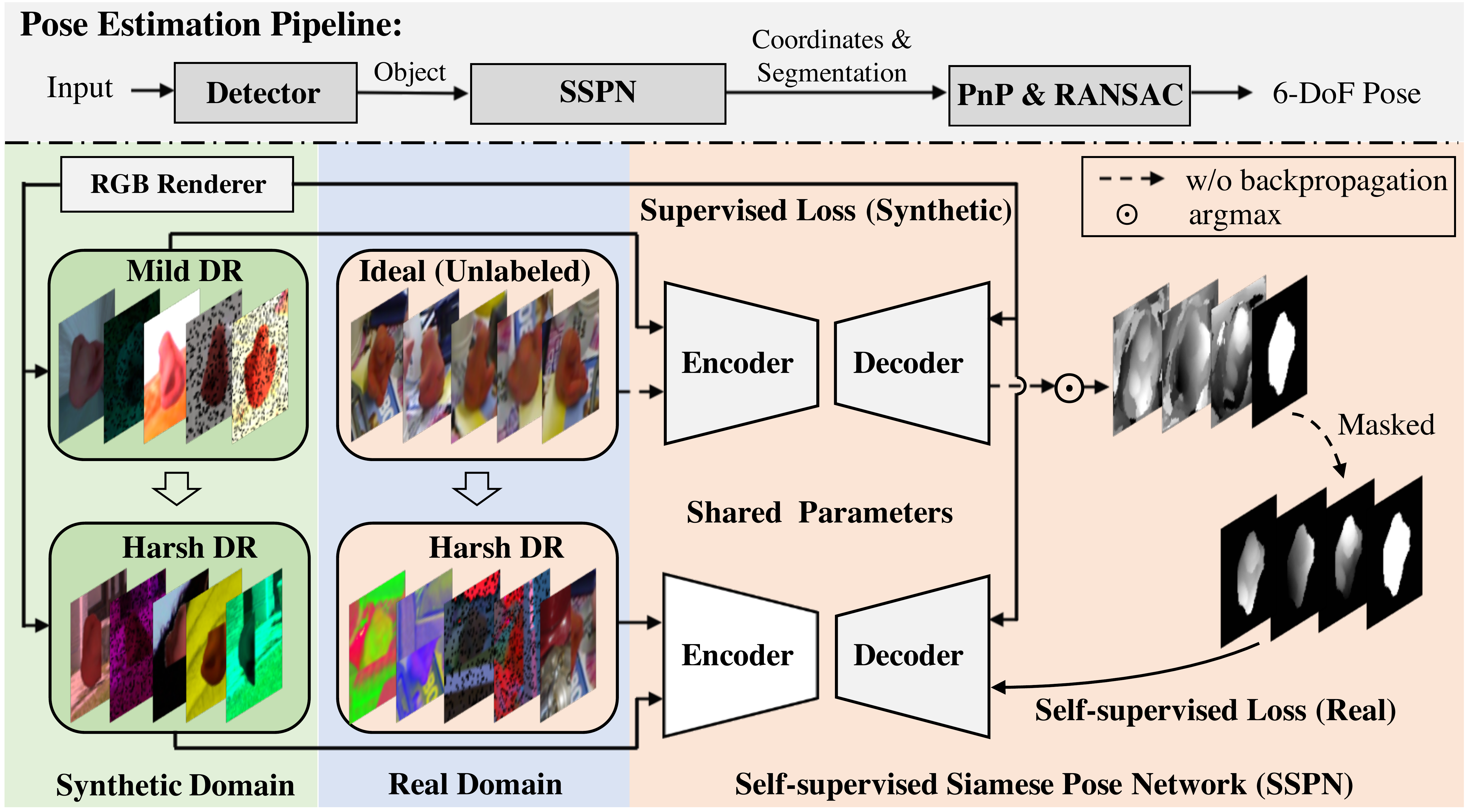}
\caption{Overview of our self-supervised siamese pose network. During training, we introduce truncation as a challenging DR to mimic severe occlusions, and utilize  unlabeled real data to bridge the domain gap and improve robustness in the real domain.}
\label{fig:sspn}
\vspace{-1em}
\end{figure}

\subsection{Self-supervised Coordinates-based Pose Estimation}
\label{sec:self_sup}
We exploit our self-supervised learning method within a coordinate-based 6-DoF pose estimation approach. Specifically, this approach uses a deep network to estimate 2D-3D correspondences by predicting the 3D location of the objects' pixels. In this section, we describe the different aspects of the overall framework.

\noindent \textbf{Locating the object in 2D.}
To build 2D-3D correspondences, we first need to identify the pixels that belong to the object. Following~\cite{park2019pix2pose,li2019cdpn}, we use a `global detection \& local segmentation' approach, leading to an efficient and flexible pose estimation framework. Specifically, we use off-the-shelf detectors (Faster-RCNN\cite{ren2015faster}, YOLOv3\cite{redmon2018yolov3}) and, for efficiency, perform `local segmentation' jointly with coordinate prediction, as discussed below.

\noindent \textbf{Self-supervised Siamese Pose Network (SSPN).}
To exploit our self-supervised loss, we introduce a Self-supervised Siamese Pose Networks (SSPN). It contains a pair of networks that share the same architecture and parameters but have their own distinct inputs. Given an unlabeled real input $X_{real}$, we perform domain randomization to obtain its counterpart
\begin{align}\label{Eq:dr}
 X^{'}_{real} =\mathcal{A}_i(\mathcal{A}_r(X))\;.
\end{align}
Then, $X_{real}$ and $X^{'}_{real}$ are fed each to one sub-network of the SSPN. Each sub-network outputs a 3D coordinate map and a segmentation mask for its input. Since the real pose annotation is unknown, to achieve self-supervised learning, each sub-network is guided to learn from the prediction of the other one.

The prediction of SSPN is a pose-relevant intermediate representation $\mathcal{O}$ (coordinates and segmentation) instead of directly the pose. 
In this case, the domain randomization operations that are relevant ($\mathcal{A}_r$) and irrelevant ($\mathcal{A}_i$) to the output can be categorized into four types: I. $\mathcal{A}_r$ (e.g., rotate, translate) that alters the pose and $\mathcal{O}$ simultaneously and consistently, where the inconsistency between $\mathcal{A}_r(f_\theta(X))$ and $f_\theta(\mathcal{A}_r(X))$ can be leveraged to provide supervision; II. $\mathcal{A}_r$ that alters the object pose but maintains $\mathcal{O}$, which yields pose ambiguities and should thus be avoided since multiple distinct poses correspond to the same representation; III. $\mathcal{A}_i$ (e.g., varying lighting, blur) that maintains both the pose and $\mathcal{O}$, where the inconsistency between $f_\theta(\mathcal{A}_i(X))$ and $f_\theta(X)$ can provide supervisory signal;  IV. $\mathcal{A}_i$ (e.g., truncation, block) that preserves the pose but alters $\mathcal{O}$, which complicates self-supervised learning since the ground truth of $f_\theta(\mathcal{A}_i(X))$ and $f_\theta(X)$ differ whereas $\mathcal{A}_i(X)$ and $X$ share the same pose. Below, we introduce our approach to exploiting self-supervised learning for coordinate and segmentation estimation.

\noindent \textbf{Coordinate estimation.}
In essence, the coordinates predicted by each of the sub-networks are supervised by those output by the other. To this end, for any of the four types of DR transformations discussed above, we can make use of the loss of Eq.~\ref{Eq:s}.
For coordinates, we compute the loss on the foreground region, since only these coordinates are effective for pose estimation. 
However, when using occlusion operations (e.g., truncation, block), the input to both branches of SSPN are inconsistent: A portion of the object that is visible in the non-DR input is hidden in the DR one.
This can be taken into account by either ignoring the occluded area in the loss, or by also penalizing discrepancies in this area. For the former case, we introduce the Masked Coordinates Visible (MCV) loss
\begin{align}
\ell_{ss}^{MCV}  =  \sum_{k=1}^{c} \sum_{j=1}^{m} \sum_{i=1}^{n} \mathbb{I}\{(i, j) \in \mathcal{A}_i(\mathcal{M})\} \cdot CE(\mathcal{C}(X)_{i, j, k}, \tilde{\mathcal{C}}(X)_{i, j, k})\;,  \label{Eq:L_MCV}
\end{align}
where $\mathcal{C}_{i, j, k}$ and $\tilde{\mathcal{C}}_{i, j, k}$ is the $k$-th coordinate predicted by the non-DR and DR sub-network at location ($i$, $j$), CE is the cross-entropy loss, and $\mathbb{I}$ is the indicator function. $\mathcal{M}$ is the non-occluded object foreground mask, and $c$, $m$, $n$ represent the number of channels, height and width of the coordinate map. 
By contrast, to enforce consistency in the entire object area, we rely on the Masked Coordinates Inpainting (MCI) loss expressed as
\begin{align}
\ell_{ss}^{MCI}  =  \sum_{k=1}^{c} \sum_{j=1}^{m} \sum_{i=1}^{n} \mathbb{I}\{(i, j) \in \mathcal{M}\} \cdot CE(\mathcal{C}(X)_{i, j, k}, \tilde{\mathcal{C}}(X)_{i, j, k})\;.  \label{Eq:L_MCI}
\end{align}
Intuitively, the MCI loss has the advantage of not biasing the output of the non-DR sub-network to the potentially-incorrect predictions of the DR ones, due to its lack of observations. By contrast, the MCV loss tends to be small on severely occluded objects because their foreground region is much smaller than that of their non-occluded counterpart. This may make the network neglect the occluded samples during training, increasing its sensitivity to occlusion. In our experiments, we will compare the effectiveness of these two loss functions.

\noindent \textbf{Segmentation loss.} 
For segmentation, we found the discrepancy between non-occluded and occluded objects not to be significant. Therefore, we simply use a weighted segmentation loss for self-supervised learning, defined as
\begin{align}
\ell^{seg}_{ss}  =  \sum_{j=1}^{m} \sum_{i=1}^{n} CE(\tilde{\mathcal{M}}(\mathcal{A}_i(X))_{i, j}, \mathcal{A}_i(\mathcal{M}(X))_{i, j})\;, \label{Eq:L_seg}
\end{align}
where $\tilde{\mathcal{M}}(\mathcal{A}_i(X))$ and $\mathcal{M}(X)$ indicate the segmentation mask obtained from the the DR-based and non-DR-based sub-network, respectively.

\noindent \textbf{Architecture.} 
For each sub-network, we employ a 34-layers ResNet as encoder to extract features, followed by a decoder consisting of bilinear-upsampling and convolutional layers to scale up and process the features used to estimate the coordinates and the object foreground region. In contrast to~\cite{park2019pix2pose,li2019cdpn}, we train SSPN in a classification manner for both coordinate estimation and segmentation, because classification was observed to be more robust than regression~\cite{wang2019NOCS}, especially in the synthetic-only case. The outputs of each sub-network are a coordinate map with $3K$ channels, where $K$ is the number of classification bins for each coordinate axis, and a 2-channel segmentation map. 

\subsection{Multi-step Self-supervised Hybrid Training}
Although our self-supervised training mechanism allows us to learn the network weights on real data without pose annotations, training solely on unlabeled real data makes the parameters drift to undesirable configurations. Intuitively, it can lead to solutions where the predicted poses are consistent but wrong.
To overcome this, we introduce the following multi-step self-supervised hybrid training.

\noindent \textbf{Step 1. Initialization.} We use parameters trained on ImageNet for initialization, as commonly done in many tasks for better learning ability.

\noindent \textbf{Step 2. Synthetic-only training.} 
This step aims to endow the model with a basic pose estimation ability. 
In the synthetic domain, since ground-truth annotations are easy to obtain, we rely on supervised learning, and
use DR to improve the domain transfer ability. We define our training loss as
\begin{align}
  \ell_{syn} = \sum_{j=1}^{m} \sum_{i=1}^{n} &(\mathbb{I}\{(i, j) \in \mathcal{M}_{syn}\} \cdot \sum_{k=1}^{c} CE(\mathcal{C}_{i, j, k}, \tilde{\mathcal{C}}(X)_{i, j, k}) + \\
   \tau\cdot & CE(\mathcal{M}_{i, j}, \tilde{\mathcal{M}}(X)_{i, j}))\;
\end{align} \label{eq:loss_syn}
where $\mathcal{M}_{syn}$ is the foreground mask in synthetic input.  $\mathcal{C}$, $\mathcal{M}$ and $\tilde{\mathcal{C}}$, $\tilde{\mathcal{M}}$ indicate the ground-truth and predicted coordinates map and segmentation map, respectively. $\tau$ is a weight to balance the terms.

\noindent \textbf{Step 3. Synthetic-real hybrid training.} 
In this step, we fuse the annotated synthetic data with the unlabeled real data to train the model in a hybrid manner.  Concretely, the synthetic and real data are sampled at the ratio 1:1 to constitute a batch to feed to the model. We use the supervised loss of Eq.~\ref{eq:loss_syn} for the synthetic data, and the self-supervised losses discussed in Section~\ref{sec:self_sup}
for the real images. The supervised loss from the labeled synthetic data endows the model with the ability to perform well on easy real samples, and the self-supervised losses from the unlabeled real images help it to work well on the hard real samples. Altogether, we write our hybrid training loss as
 \begin{align}
   \ell  =  \ell_{syn}  + \beta \cdot \ell^{coor}_{ss}  +  \gamma \cdot \ell^{seg}_{ss}\;,
  \label{eq:loss_hyb}
\end{align}
where $\ell_{syn}$ represents the supervised loss for synthetic input. $\ell^{coor}_{ss}$and $\ell^{seg}_{ss}$ are self-supervised losses for coordinates and segmentation for real input. $\beta$ and $\gamma$ are weights for balance each term. 

Our self-supervised approach has the following merits: 1) It makes the model trainable on real-world data without requiring any real 6-DoF annotation; 2) Self-supervised training significantly improves the model's robustness in the real-world domain; 3) It influences only the training process and brings no burden during inference. 4) It generalizes to a variety of pose estimation approaches. 

\subsection{Generalization}
Let us now briefly explain how our self-supervised training method generalizes to other pose estimation frameworks. 
For the direct pose estimation techniques, such as~\cite{kehl2017ssd}, which directly estimate the pose from the image, since the truncation and block operations do not influence the prediction, the output from the non-occluded input directly provides supervision for the truncated input. For keypoints-based methods, such as~\cite{hu2019segmentation,peng2019pvnet}, where the model is trained to predict a keypoint-relevant representation, self-supervision can be applied using the $MCI$ or $MCV$ loss, as in our approach.

\section{Experiments}

\subsection{Experimental Setup\label{Sec.Details}}
\noindent \textbf{Datasets.}
In addition to our own Rand-LINEMOD dataset, described in Section~\ref{sec:DR}, we conduct experiments on the LINEMOD, Occluded LINEMOD and YCB datasets. The LINEMOD dataset contains 15 poorly-textured objects imaged in cluttered scenes without occlusion, and is the \emph{de facto} standard benchmark for poorly-textured object pose estimation. We follow~\cite{li2018deepim} to split the dataset. Occluded LINEMOD~\cite{brachmann2016uncertainty} was collected by annotating eight objects with heavy occlusions in a video sequence of LINEMOD. It is widely used to evaluate the pose estimation performance on occluded objects. The YCB dataset contains 21 gadgets observed in 92 sequences. For these datasets, we render 10000 images for each category using Blender, following to the pose distribution of the training set, and use images from the PASCAL VOC2012~\cite{VOC} and MS COCO~\cite{lin2014microsoft} datasets as background images. Furthermore, for all experiments in the self-supervised setting, we discard the pose annotations provided with the real training images. To obtain 2D detections, we trained FasterRCNN~\cite{ren2015faster} and YOLOv3~\cite{redmon2018yolov3} on the training set of the LINEMOD and Occluded LINEMOD datasets. 

\noindent \textbf{Implementation details.}
We set the number of bins in the output coordinate space to $K=65$, where 64 are valid coordinate bins and the additional one accounts for the background. The resolutions of the input and output are 256$\times$256 and 64$\times$64, respectively. The parameters $\tau$, $\beta$, $\gamma$ in the loss are set to 1, 1, 0.01, respectively. See the supplementary material for the training details. 

\noindent \textbf{Metrics.}
To evaluate performance, we report four common metrics:  5cm\ 5$^\circ$, Proj. 2D, ADD and ADD-AUC. 
For 5cm\ 5$^\circ$, a pose is correct if the translation error and rotation error are smaller than $5cm$ and $5^\circ$, respectively.
For Proj. 2D, the estimated pose is correct if the average 2D projection error is smaller than 5 pixels.
For ADD, a pose is deemed correct if the average vertex distance in 3D space is below 0.1$d$, where $d$ is the object diameter. For symmetric objects, the nearest points are used to compute the distance.
ADD-AUC is obtained by varying the threshold in ADD and computing the area under the resulting curve. The details of these metrics can be found in the supplementary material.

\setlength{\tabcolsep}{6pt}
\begin{table}[t]
\tabcolsep=0.1cm
\small
\begin{center}
\caption{Comparison of the MCV and MCI losses without and with self-supervised real data on the Occluded LINEMOD dataset. (Metric: ADD-0.1d; Using ground-truth bounding boxes; {\bf T} indicates truncation DR and  \textbf{S}  self-supervised learning.)}
\vspace{-0.5em}
\label{table:MCI_MCV}
\begin{tabular}{c|cccccccc|c}
\hline
Object       & Ape  & Can   &   Cat   &   Dril.   & Duck   &   Eggb.   &   Glue   &   Hol.  &   Avg \\
\hline
w/o \textbf{T} \& MCV loss
		& 21.4  & 18.1   & 15.8   & 14.0   & 26.1    & 37.8   & 33.1      & 27.5   & 24.2 \\
 \textbf{T}  \& MCV loss
		& 23.9 &  56.2  & 18.5   & 41.2   & 28.84     &  40.3    & 43.3   & 21.2  & 34.2   \\
 \textbf{T}  \& MCI loss 
		&  23.1  & 64.4   &  17.4   &43.7   &  34.4    & 35.7     &  43.2   &  23.8  &    35.7 \\
 \textbf{T}  \& MCI loss \& \textbf{S}
		& \textbf{29.4}  & \textbf{74.0}   &  \textbf{23.8}   & \textbf{53.3}    &  \textbf{33.6}     & \textbf{50.9}      & \textbf{56.5}     & \textbf{26.5}   &   \textbf{43.5} \\
\hline
\end{tabular}
\vspace{-2em}
\end{center}
\end{table}

\subsection{Ablation Study}

\noindent \textbf{{MCV loss vs. MCI loss.}}
We compare the MCV and MCI losses on the Occluded LINEMOD dataset. Specifically, we evaluate their use on synthetic data only. Concretely, we use the model trained with various DR but \textit{without} truncation as a baseline. Then, we introduce the truncation to the synthetic input and use the MCV loss and MCI loss respectively in training.
As shown in Table~\ref{table:MCI_MCV}, in the synthetic-only case, using the MCV loss with additional truncation DR improves the performance from 24.2\% to 34.2\%, but is outperformed by using the MCI loss. Employing this loss in conjunction with self-supervised real data yields a further significant accuracy boost to 42.3\%. Hereinafter, we adopt the MCI loss in self-supervised training.

\noindent \textbf{{Self-supervised learning}.}
We further evaluate the importance of self-supervised learning on the LINEMOD, Rand-LINEMOD and YCB datasets. For comparison, we use the backbone network trained with synthetic data and various DR strategies, \textit{including} truncation as the baseline. To focus on the pose estimation, here we use the ground-truth bounding boxes for detection.

\setlength{\tabcolsep}{6pt}
\begin{table}[tb]
\begin{center}
	\small
	\caption{\label{table:ab_LM} Ablation study of self-supervised learning on LINEMOD. (Note: The ground-truth bounding boxes are used. \textbf{S} is our self-supervised learning.) }
    \vspace{-0.5em}
	\begin{tabular}{c|c|c|c|c|c|c}
\hline
    Metric			& \multicolumn{2}{c|}{ADD}       & \multicolumn{2}{c|}{Proj. 2D}   & \multicolumn{2}{c}{5cm\ 5$^\circ$}  \\ \hline
    Method  		& LM  	& LM(\textbf{S})         & LM  	& LM(\textbf{S})  		& LM  	& LM(\textbf{S})	 \\ 
\hline
    Accuracy		& 67.6 	 & \textbf{81.1} 		   &87.1	&\textbf{90.9}			&  74.7	  & \textbf{85.7}		 \\ 
\hline
\end{tabular}
\vspace{-1em}
\end{center}
\end{table}

\setlength{\tabcolsep}{6pt}
\renewcommand{\arraystretch}{1.1}
\begin{table}[tb]
\begin{center}
	\small
	\caption{\label{table:ab_RLM} Effect of self-supervised learning on Rand-LINEMOD (RLM). (Note: The ground-truth bounding boxes are used.  \textbf{S} is our self-supervised learning.) }
    \vspace{-0.5em}
	\begin{tabular}{c|c|c|c|c|c|c}
\hline
    Metric			& \multicolumn{2}{c|}{ADD}       & \multicolumn{2}{c|}{Proj. 2D}   & \multicolumn{2}{c}{5cm\ 5$^\circ$}  \\ \hline
    Method  	 	& RLM   	& RLM(\textbf{S})    & RLM   	& RLM(\textbf{S})  & RLM   	& RLM(\textbf{S})  \\
\hline
    Blur	   			& 67.8 	 & \textbf{80.7}	  & 87.1	&  \textbf{90.7}	&  75.1 & \textbf{85.4}	   \\ \hline
    Contrast		& 63.9   & \textbf{78.9}   &84.0 	&\textbf{90.0}	& 70.5  &  \textbf{83.4} 	\\ \hline
    Lighting		& 65.2 	 & \textbf{79.6}	  & 85.5	&  \textbf{89.8	}&  72.8	 & \textbf{83.8}		\\ \hline
    Block			   & 66.7 	& \textbf{78.6}	  & 85.8	& \textbf{90.1}	&  72.8	&\textbf{82.5}	   \\ \hline
    Truncation	   & 31.5 	& \textbf{47.4}	  & 47.3	&  \textbf{68.8} &  26.4	& \textbf{48.1}	   \\ 
\hline
 \end{tabular}
 \vspace{-2em}
\end{center}
\end{table}

\setlength{\tabcolsep}{6pt}
\renewcommand{\arraystretch}{1.1}
\begin{table}[tb]
\begin{center}
	\small
	\caption{\label{table:ab_YCB} Ablation study of self-supervised learning on the YCB dataset. (Note: The ground-truth bounding boxes are used. \textbf{S} is our self-supervised learning.) }
    \vspace{-0.5em}
	\begin{tabular}{c|c|c|c|c|c|c}
	\hline
    Metric			& \multicolumn{2}{c|}{ADD-AUC}       & \multicolumn{2}{c|}{Proj. 2D}   & \multicolumn{2}{c}{5cm\ 5$^\circ$}  \\  \hline
    Method  		& YCB  	& YCB(\textbf{S})         & YCB  	& YCB(\textbf{S})  		& YCB  	& YCB(\textbf{S})	 \\ \hline
    Accuracy		& 46.6 	 & \textbf{50.5} 		   		  & 13.5	    &\textbf{15.6}			     &  20.6	  & \textbf{24.7}		 \\ \hline
    	\end{tabular}
\end{center}
\end{table}

\setlength{\tabcolsep}{2pt}
\begin{table}[tb]
\tabcolsep=0.065cm
\scriptsize
\begin{center}
\caption{Comparison with the state-of-the-art RGB-only approaches on LINEMOD. (Metric: ADD. \textbf{Y} and \textbf{F} indicate YOLOv3\cite{redmon2018yolov3} and Faster-RCNN\cite{ren2015faster} for detection. \textbf{S} is our self-supervised learning.)}
\vspace{-0.5em}
\label{table:lm_sota}
\begin{tabular}{c|ccccccccccccc|c}
\hline
Object       & Ape   & Bv.   & Cam.   & Can   &   Cat   &   Dril.   & Duck   &   Eggb.   &   Glue   &   Hol.   &   Iron   & Lamp   & Ph.   &   Avg \\
\hline
SSD6D\cite{kehl2017ssd} 	
                 & 2.6    & 15.1   & 6.1   & 27.3   & 9.3    & 12.0   & 1.3       & 2.8   & 3.4   & 3.1    & 14.6      & 11.4& 9.7  & 9.1 \\
AAE\cite{AAE_2018_ECCV} 		
                 & 4.0    & 20.9   & 30.5   & 35.9   & 17.9    & 24.0    & 4.9       & 81.0   & 45.5   & 17.6    & 32.0      & 60.5   & 33.8  & 31.4 \\
DPOD\cite{zakharov2019dpod}
	 	& 37.2  & 66.8   & 24.2   & 52.6   & 32.4   &  66.6    & 26.1     & 73.4   & 75.0  &  24.5   &  85.0      & 57.3   & 29.1  &  50.0 \\
\textbf{Ours(Y)}
 		& 57.1 & 85.7   & 47.3   & 88.6   &  72.0    & 68.1    & 44.1   & 97.3  &   89.0   &   22.7   &   76.5   & 78.9   & 44.0   & 67.0\\
 \textbf{Ours(F)}
 		& 56.5 & 87.0   & 48.2  & 86.8   &  73.3    & 66.1    & 44.7   & 98.0  &   89.9   &   22.6   &   75.7   & 81.6   & 41.6   & 67.1\\
\textbf{Ours(S\&Y)}
		& 64.9  & 93.0    &\textbf{ 85.9}  & 98.4  & 87.3     &  91.0    & 61.4    & 94.7   & \textbf{92.0}  &  28.0   &  88.2      & 92.1   & 67.0  &  80.3 \\      
\textbf{Ours(S\&F)}
		& \textbf{64.1}  & \textbf{93.4}    & \textbf{85.9}  & \textbf{97.4}  & \textbf{86.7}     &  \textbf{90.8}    & \textbf{61.7}    & \textbf{94.9}   & \textbf{92.0}  &  \textbf{31.1}   &  \textbf{87.7}      & \textbf{93.6}   & \textbf{66.2}  &  \textbf{80.4} \\   
\hline
BB8\cite{rad2017bb8}		
                & 27.9  & 62.0  & 40.1    & 48.1   & 45.2     & 58.6    & 32.8    & 40.0   & 27.0   & 42.4    & 67.0      & 39.9  & 35.2  &  43.6 \\  
YOLO6D\cite{tekin18_yolo6d}  
                 & 21.6  & 81.8  & 36.6    & 68.8   & 41.8     & 63.5    & 27.2    & 69.6   & 80.0   & 42.6    & 75.0      & 71.1  & 47.7  &  56.0  \\
PoseCNN\cite{xiang2018posecnn} 
                 &  27.8 & 68.9  & 47.5    & 71.4   & 56.7     & 65.4    & 42.8    & 98.3   & 95.6   & 50.9     & 65.6      & 70.3  & 54.6 &  62.7 \\
Pix2Pose\cite{park2019pix2pose} 
                 & 58.1  & 91.0  & 60.9    & 84.4   & 65.0      & 76.3   & 43.8     & 96.8 & 79.4     & 74.8    & 83.4      & 82.0  & 45.0  & 72.4 \\
PVNet\cite{peng2019pvnet}
		& 43.6   & 99.9  & 86.9   & 95.5   & 79.3   &   96.4   & 52.6   &   99.2   & 95.7   &  81.9   &   98.9   & 99.3   & 92.4   &   86.3 \\
CDPN\cite{li2019cdpn}
		& 64.4   & 97.8  & 91.7   & 95.89  & 83.8   &   96.2   & 66.8   &   99.7   & 99.6   &  85.8   &   97.9   & 97.9   & 90.8   &  89.9 \\
\hline
\end{tabular}
 \vspace{-1em}
\end{center}
\end{table}
\setlength{\tabcolsep}{1.4pt}

\setlength{\tabcolsep}{6pt}
\begin{table}[tb]
\tabcolsep=0.1cm
\small
\begin{center}
\caption{Comparison with the state-of-the-art RGB-only methods on Occluded LINEMOD. (Metric: ADD. \textbf{F} is Faster-RCNN, \textbf{T} and \textbf{S} denote the proposed truncation operation and self-supervised learning. \textbf{R} represents using labeled real data.)}
\vspace{-0.5em}
\label{table:lmo_sota}
\begin{tabular}{c|ccccccccccccc|c}
\hline
Object       & Ape  & Can   &   Cat   &   Dril.   & Duck   &   Eggb.   &   Glue   &   Hol.  &   Avg \\
\hline
YOLO6D\cite{tekin18_yolo6d}  	
                 & 2.48 & 17.48  & 0.67   & 7.66   & 1.14    & -         & 10.08     & 5.45   & 6.42 \\
PoseCNN\cite{xiang2018posecnn} 	
                 & 9.6   & 45.2    & 0.93   & 41.4   & 19.6    & 22.0   & 38.5       & 22.1   & 24.9 \\
Oberweger\cite{oberweger2018making}
		& 17.6  & 53.9   & 3.31   & 62.4   & 19.2    & 25.9   & 39.6      & 21.3   & 30.4  \\ 
PVNet\cite{peng2019pvnet}
		& 15.8  & 63.3   & 16.7   & 25.2   & 65.7    & 50.2   & 49.6      & 39.7   & 40.8  \\
\textbf{Ours(T \& F)}
		& 23.1  & 64.4   & 17.4   & 43.7   &  34.4    & 35.7     & 43.2   & 23.8  &   35.7 \\
\textbf{Ours(TS \& F)}
		& \textbf{28.6}  & \textbf{72.9}   &  \textbf{23.3}   & \textbf{52.3}   &  \textbf{35.8}    & \textbf{46.9}     &  \textbf{50.5}   & \textbf{28.1}  &    \textbf{42.3} \\
\hline
\textbf{Ours(R \& F)} & \textit{31.4}  & \textit{53.3}   & \textit{17.1}   & \textit{31.7}   &  \textit{34.6}    & \textit{43.1}     & \textit{39.4}   & \textit{43.1}  &   \textit{36.7} \\
\textbf{Ours(TR\&F)} & \textbf{\textit{34.2}}  & \textit{\textbf{72.9}}   & \textit{\textbf{26.7}}   & \textit{\textbf{66.8}}   &  \textit{\textbf{40.0}}    & \textit{\textbf{60.4}}     & \textit{\textbf{52.6}}   & \textit{\textbf{57.0}}  &   \textit{\textbf{51.3}} \\
\hline
\end{tabular}
 \vspace{-1em}
\end{center}
\end{table}

On LINEMOD, as can be seen in Table~\ref{table:ab_LM}, our self-supervised learning scheme improves all metrics (see the detailed results in the supplementary material). The accuracy is increased from 67.6\% to 81.06\% on ADD, from 78.1\% to 90.9\% on Proj. 2D and from 74.7\% to 5.7\% on  5cm\ 5$^\circ$. This clearly evidences the effectiveness of our approach at leveraging unlabeled real data to assist training in the real domain, even without heavy occlusions. 

On Rand-LINEMOD, as shown in Table~\ref{table:ab_RLM}, our approach improves the performance of the baseline on all test sequences and on all metrics (see detailed results in the supplementary material). In particular, with truncation DR, the performance improves from 31.5\% to 47.4\% on ADD, from 47.3\% to 68.8\% on Proj. 2D and from 26.4\% to 48.1\% on 5cm\ 5$^\circ$. This demonstrates the effectiveness of self-supervised learning at improving robustness to heavy occlusions. 

In the YCB dataset, many training samples are heavily occluded, blurred, etc., which influences the effect of our self-supervised learning. However, as shown in Table~\ref{table:ab_YCB}, our approach still achieves generally better results than the baseline.

\subsection{Comparison with State-of-the-art Approaches}

\noindent \textbf{{LINEMOD dataset}.}
We first compare our approach with state-of-the-art techniques on the LINEMOD dataset. Some of the methods (SSD6D~\cite{kehl2017ssd}, AAE~\cite{AAE_2018_ECCV}, DPOD\cite{zakharov2019dpod}) focus on pose estimation without real pose annotations. Specifically, SSD6D trains a rotation classifier based on SSD~\cite{liu2016ssd}, and AAE trains an augmented autoencoder with various DR for pose retrieval. These approaches attempt to directly estimate or retrieve the rotation from the image, and their inferior performance evidences the challenging nature of this task. By contrast, DPOD follows a coordinate-based approach but in a semantic segmentation framework. It constitutes the state of the art for pose estimation without real pose annotations. As shown in Table~\ref{table:lm_sota}, however, our SSPN with DR outperforms DPOD by a significant margin. Specifically, with self-supervision, we reach the state-of-the-art accuracy of 80.4\%. Note that our approach even surpasses many methods (e.g., Pix2Pose~\cite{park2019pix2pose}, YOLO6D~\cite{tekin18_yolo6d}, PoseCNN~\cite{xiang2018posecnn},~\cite{brachmann2016uncertainty}) that rely on annotated real data during training, and approaches the state-of-the-art accuracy of 89.9\% obtained by CDPN~\cite{li2019cdpn}.

\noindent \textbf{{Occluded LINEMOD dataset}.}
We then compare our approach with the state-of-the-art techniques on the challenging Occluded LINEMOD dataset. As shown in Table~\ref{table:lmo_sota}, training our SSPN with DR and self-supervised learning yields an accuracy of 42.3\%, outperfoming the state-of-the-art accuracy of 40.8\% obtained by PVNet. The experiments show that our self-supervised learning strategy significantly helps pose estimation without requiring real pose annotations. For reference, we also evaluate SSPN when annotated real data is also involved. This further improves the performance to 51.3\%, which constitutes the state of the art on this dataset, outperforming the competitors by a large margin.

\begin{figure}[t]
\centering
\includegraphics[width=12cm]{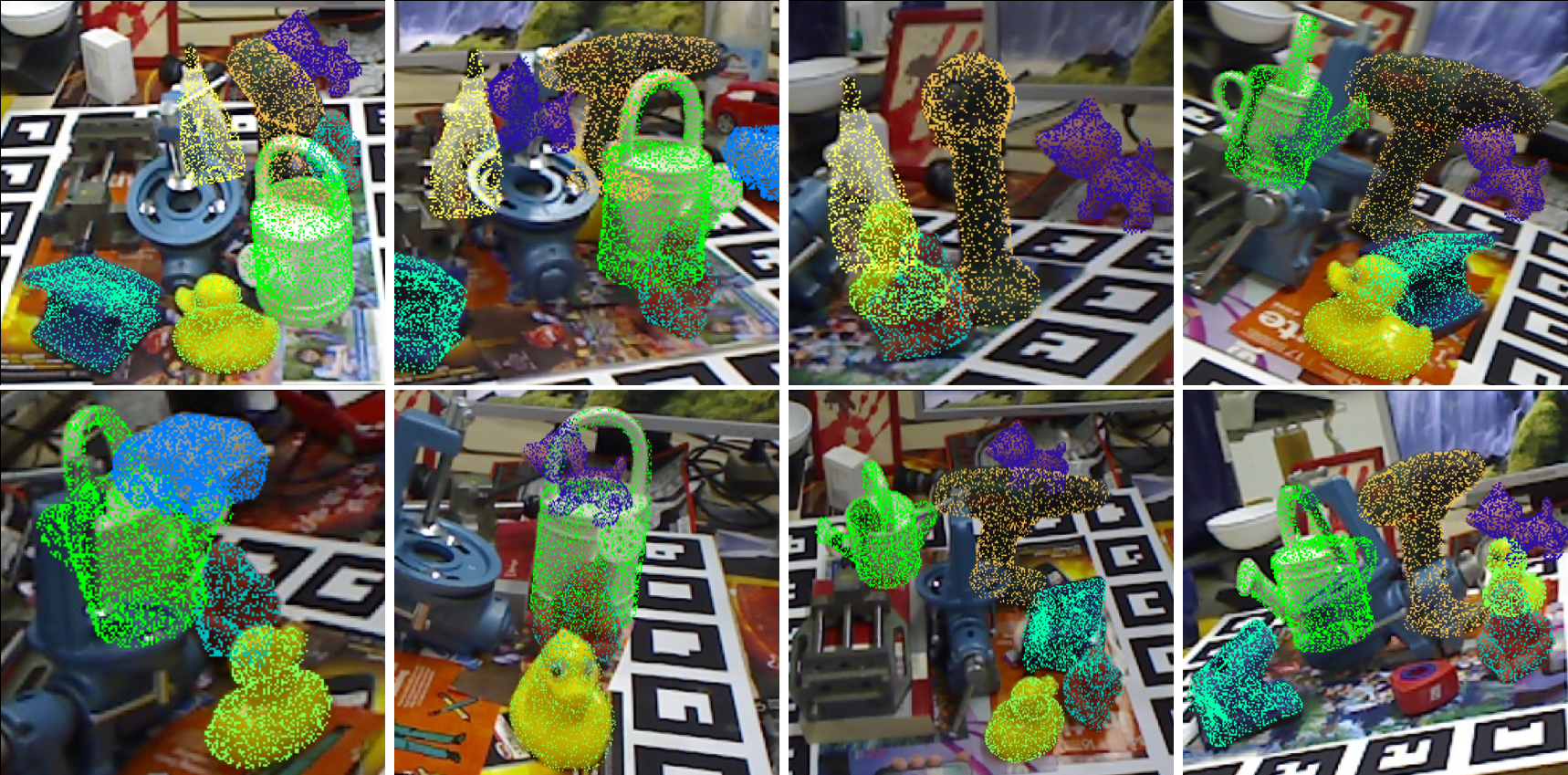}
\caption{Qualitative results on Occluded LINEMOD with severe occlusions. We show the object model projected to the image with the estimated pose. Self-supervised training with truncation makes our model robust to severe occlusions. More results can be found in the supplementary material.}
\label{fig:show}
\vspace{-0.5em}
\end{figure}

\section{Conclusions}
We have proposed a self-supervised learning approach for 6-DoF object pose estimation that enables the network to leverage real data without pose annotations during training. Our comprehensive experiments have shown that our approach significantly improves the pose estimation performance and robustness in the real domain. We achieve the state-of-the-art performance on the challenging Occluded LINEMOD dataset, and outperform the state-of-the-art synthetic-only approaches by a large margin. In the future, we intend to generalize our approach to other related tasks, such as camera localization.

\bibliographystyle{splncs04}
\bibliography{egbib}
\end{document}